# Agentic AI-enabled Semantic Commissioning of a Cognitive Digital Twin for Reconfigurable Manufacturing

Yangyang Liu, Xun Xu, Jan Polzer

*Abstract*— **Rapid bespoke commissioning of the Cognitive Digital Twin (CDT) is a major challenge in reconfigurable manufacturing. Traditional digital twin (DT) construction methods primarily focus on geometric reconstruction, often neglecting the deep semantic integration and functional interoperability necessary for autonomous reasoning. This paper proposes an agent-based, AI-driven workflow to automate end-to-end CDT debugging. The system utilises LangGraph as a multi-agent orchestration engine to achieve dual-path synthesis: the semantic path extracts technical specifications from unstructured documents using Retrieval Augmented Generation (RAG), while the functional path autonomously discovers and binds to real-time industrial telemetry data using Model Context Protocol (MCP). Experimental validation in a robotic machining cell demonstrates that the system achieves a mean average accuracy (mAP) of 97.2% in perception and reduces the deployment cycle from several weeks to an average of 2 hours, marking a paradigm shift from manual scripting to autonomous orchestration.**

## I. Introduction

CDT extends traditional DT by integrating advanced reasoning and autonomous learning [1]. Unlike standard DTs that primarily serve as passive data reflections, a CDT is expected to perceive physical changes and autonomously adapt its internal models [2]. However, a persistent "Commissioning Bottleneck" impedes the agility of modern production lines. As leading automotive manufacturers transition toward high-variance, low-volume (HMLV) production [3], driven by electrification and mass customisation, the frequency of workshop reconfiguration has scaled exponentially [4]. This shift necessitates Reconfigurable Manufacturing Systems (RMS) and modular workstations (e.g., using AMRs and flexible framing systems) to accommodate non-standard product routings and specialised EV components like battery packs. Manual calibration of digital replicas in such highly dynamic and manual-intensive environments introduces a significant "oversight lag," [5, 6] where the virtual model fails to reflect physical reality for days or weeks, rendering real-time optimisation impossible.

Existing research in automated DT construction often focuses on geometric reconstruction [7], frequently overlooking the semantic enrichment and functional interoperability required for a true CDT [8]. A functional CDT must not only "look" like its physical counterpart but also "know" its technical specifications and "connect" to its live operational states. Bridging the gap between raw visual perception and a semantically rich digital replica requires a sophisticated orchestration of heterogeneous AI tools and industrial communication standards.

To address these challenges, this paper proposes an Agentic Workflow for the rapid semantic setup of CDT, validated within a physical robotic machining workshop. By leveraging Large Language Models (LLMs) as a reasoning core, the system autonomously orchestrates a multi-stage pipeline:

- Perception and 3D Reconstruction: The workflow begins with a deep-learning-based vision module to monitor the physical layout. Upon detecting reconfigurations, the agent retrieves corresponding 3D models from a database to reconstruct the spatial scene.

- Hybrid Semantic Enrichment: To ensure the precision of the digital replica, we employ RAG to extract static technical parameters from unstructured PDF documentation into structured property graphs, eliminating "hallucinations" regarding equipment specifications. To bridge the "Heterogeneity Gap," we implement the MCP [9, 10] as a standardised mediation layer. Unlike traditional hard-coded data bindings, MCP enables LLM agents to autonomously discover and subscribe to OPC UA server nodes by transforming complex industrial telemetries into steerable tool calls. This dual-path synthesis, combining RAG [11] for static specification extraction with MCP for dynamic state synchronization, ensures both semantic depth and functional responsiveness.

- Knowledge Integration and Visualisation: All extracted spatial, static, and dynamic data are synthesised into a KG, which serves as the structured semantic foundation of the CDT. The final synchronised scene is rendered via a Three.js web interface, providing a lightweight and cross-platform visualisation.

A key feature of our workflow is the Human-in-the-Loop (HITL) mechanism. When the agent encounters perceptual uncertainties, such as occlusions or unregistered equipment, it prompts for human intervention. This feedback is integrated into a persistent memory module, enhancing the system's future autonomy and robustness in complex industrial scenes.

The primary contributions of this work are summarised as follows:

- An Autonomous Agentic Bespoke Commissioning Framework: A systematic orchestration using LLM-based agents that transitions CDT construction from manual, static configuration to an active "Perception-Reasoning-Action" loop. In this context, the 'Action' closes the feedback loop by enabling the CDT to autonomously generate and push parameterised control signals back to the physical shop-floor via bi-directional OPC UA write-nodes.

- Empirical Validation on a Physical Testbed: A full-scale implementation in a reconfigurable laboratory workshop, proving the framework's efficacy in reducing bespoke commissioning time from hours to minutes while maintaining high semantic fidelity.

The remainder of this paper is organised as follows: Section II reviews related work; Section III details the proposed framework and its experimental validation; and Section IV concludes the paper.

## II. Related Work

### A. *Evolution and Architecture of CDT*

DT technology has undergone a significant transformation from static virtual replicas to dynamic, interactive systems. However, traditional DTs are often limited by hard-coded data bindings and lack autonomous reasoning capabilities in complex, reconfigurable environments [12]. To bridge these gaps, CDT architectures have emerged by integrating cognitive functions such as perception, reasoning, and learning [2]. As synthesised in recent surveys [2], CDT frameworks have progressed from early ontology-driven models (2020) to multi-dimensional cognitive models (from 2022 to 2024) that utilise KG for industrial logic representation.

Despite this theoretical evolution, the transition of CDT from laboratory settings to industrial shop floors faces a severe bottleneck during the bespoke commissioning phase. Sun et al. [13] highlight that current virtual commissioning technologies often suffer from a detachment from actual operation scenarios, which diminishes the overall commissioning effect. Central to this issue is the semantic mapping process, the effort to define, link, and interpret the relationships between heterogeneous physical assets (e.g., IoT sensors, machine tools) and their virtual counterparts. This process remains a largely manual, time-consuming, and labour-intensive endeavour. The bottleneck in semantic mapping is driven by four critical technical challenges identified in both literature and industrial practice: Heterogeneity of Data [14], The "Bespoke" Problem [13], Complexity of Knowledge Extraction and System Fragility and Maintenance [15].

### B. *LLM-driven Autonomous Agents in Manufacturing*

Unlike standalone Large Language Models (LLMs), agentic frameworks emphasise iterative reasoning and autonomous tool-calling to solve multi-step engineering tasks. While early research primarily focused on high-level task planning [16], recent developments in 2025 and 2026 signal a paradigm shift toward autonomous low-level execution [17].

Building upon the potential of Multi-Agent Systems (MAS) in flexible production [18], the emergence of graph-based orchestration frameworks, such as LangGraph, addresses the research gap in managing complex, stateful manufacturing workflows [17]. These frameworks enable agents to move beyond strategic planning into the direct manipulation of industrial assets. Specifically, by integrating with the MCP and Industrial IoT (IIoT) interfaces, agents can now autonomously navigate industrial protocols and interact with Programmable Logic Controllers (PLCs) without manual scripting [17].

To support such low-level interaction, understanding the underlying data structures of industrial control, such as the Sequential Function Chart (SFC) defined in the IEC61131-3 standard [19], is essential for agents to interpret and generate executable logic.

Moreover, the introduction of standardised Agent-to-Agent (A2A) protocols in 2025 facilitates interoperability in heterogeneous environments. To meet the high reliability and safety standards of smart manufacturing, current trends advocate for HITL architectures. These systems combine agentic reasoning with user-friendly interfaces to ensure transparency and auditability during the transition from simulation to real-world production deployment [16, 17].

### C. *Knowledge Representation and Standardised Interfacing*

Effective knowledge representation is the backbone of CDT. Adhering to international standards is crucial for interoperability. Current research emphasises the use of ISO 23247 [20] for defining entity attributes and RAMI 4.0 (ISO 3290) [21] for structural alignment within the administrative shell.

To bridge the "Semantic Gap" during data integration, recent approaches leverage RAG to extract specifications from unstructured documents [22]. However, a critical missing link is a standardised protocol that allows LLM-based agents to interact with diverse data sources and industrial tools seamlessly. The MCP [9] has recently emerged as an open standard to enable this integration, providing a modular SDK-based approach (e.g., Python/TypeScript) to connect LLMs with external contexts. Our work is among the first to implement MCP within a CDT framework to automate the binding of real-time industrial telemetry.

## III. Agentic Framework Realisation and Experimental Validation

This section addresses the practical implementation of the proposed cognitive-function CDT framework in a real-world–inspired robotic machining cell. We detail the instantiation of each cognitive module using an LLM-driven Agentic workflow and report the experimental results from both a perception and system-integration perspective.

### A. *Experimental Setup & Bespoke Commissioning Context*

*Physical Scenario:* The testbed is a reconfigurable discrete manufacturing machining cell consisting of three core entities (as shown in Figure 1):

- Processing Equipment: Two industrial robots (e.g., ABB and KUKA models) responsible for assembly and machining tasks.
- Material-Handling Facilities: Two automated conveyor belts (linear and serpentine types) equipped with sensors for material flow.
- Monitoring Device: An RGB-D depth camera (Intel RealSense D435i) mounted above the cell to capture spatial layouts.

In this scenario, "Production Rotation" occurs frequently, where layout adjustments and equipment swaps are required to accommodate different product variants, mirroring the agile manufacturing needs of companies like BMW or Toyota.

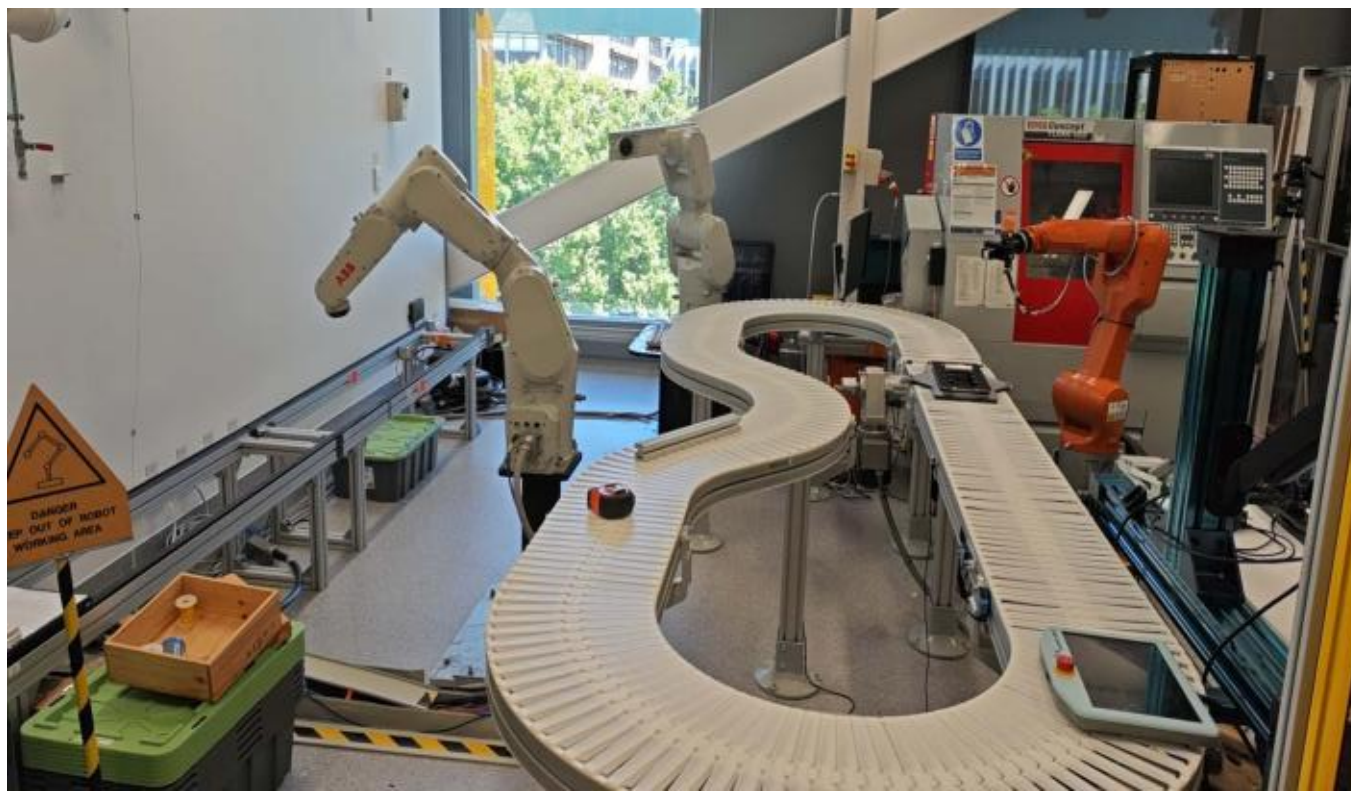

Figure 1. A modular and reconfigurable robotic machining cell for CDT validation

*Technology Stack & Semantic Foundation:* To realise the "Perception-to-Digital" closed loop, the following stack is deployed:

- Perception: YOLOv11 [23] for object detection and pose estimation.
- Cognition Core: GPT-5.1 via the LangGraph framework for multi-agent reasoning.
- Semantic Memory: Neo4j Graph Database serves as the KG backbone, storing the relationship between geometry, static specifications, and dynamic telemetry.
- Action & Visualisation: FreeCAD for 3D asset assembly and Three.js for a lightweight, web-based, synchronised monitoring interface.

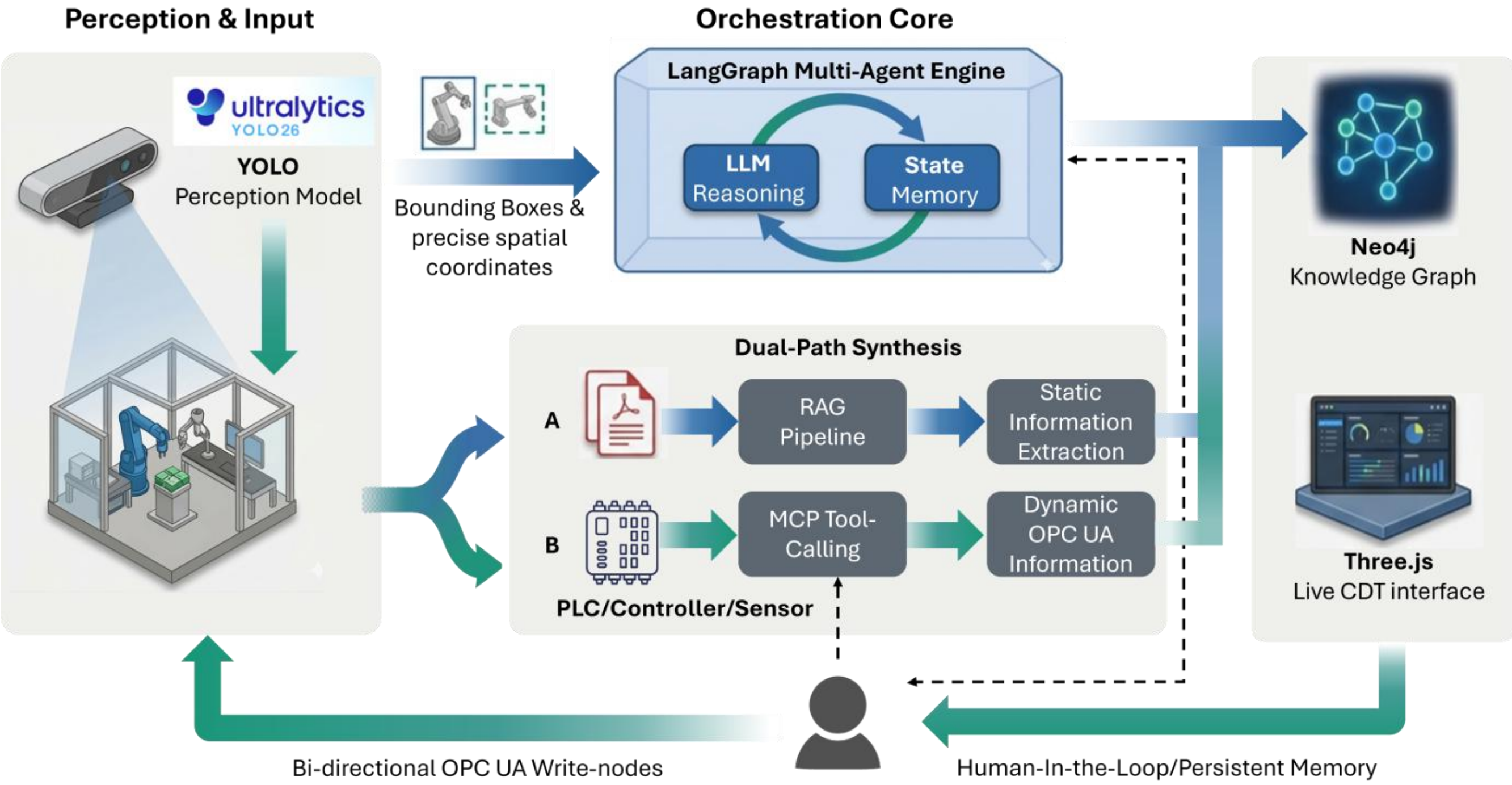

Figure 2. Agentic AI-enabled Semantic Commissework for CDT.

## B. *The Agentic Bespoke Commissioning Workflow*

To ensure a robust transition from raw perception to a live DT, the system orchestrates a multi-stage Agentic Workflow. Unlike traditional hard-coded scripts, this workflow treats bespoke commissioning as a dynamic reasoning task.

*Stage I Hybrid Semantic Enrichment:* Multi-modal Semantic Ingestion Upon detecting an equipment instance via YOLOv11, the agent triggers a MinerU-powered RAG pipeline. The agent parses unstructured PDF manuals to extract a structured attribute vector $A_i$ (e.g., Rated Capacity, MaxPayload). This links visual bounding boxes to physical capabilities, forming a semantic node in the Neo4j graph (as shown in Figure 3).

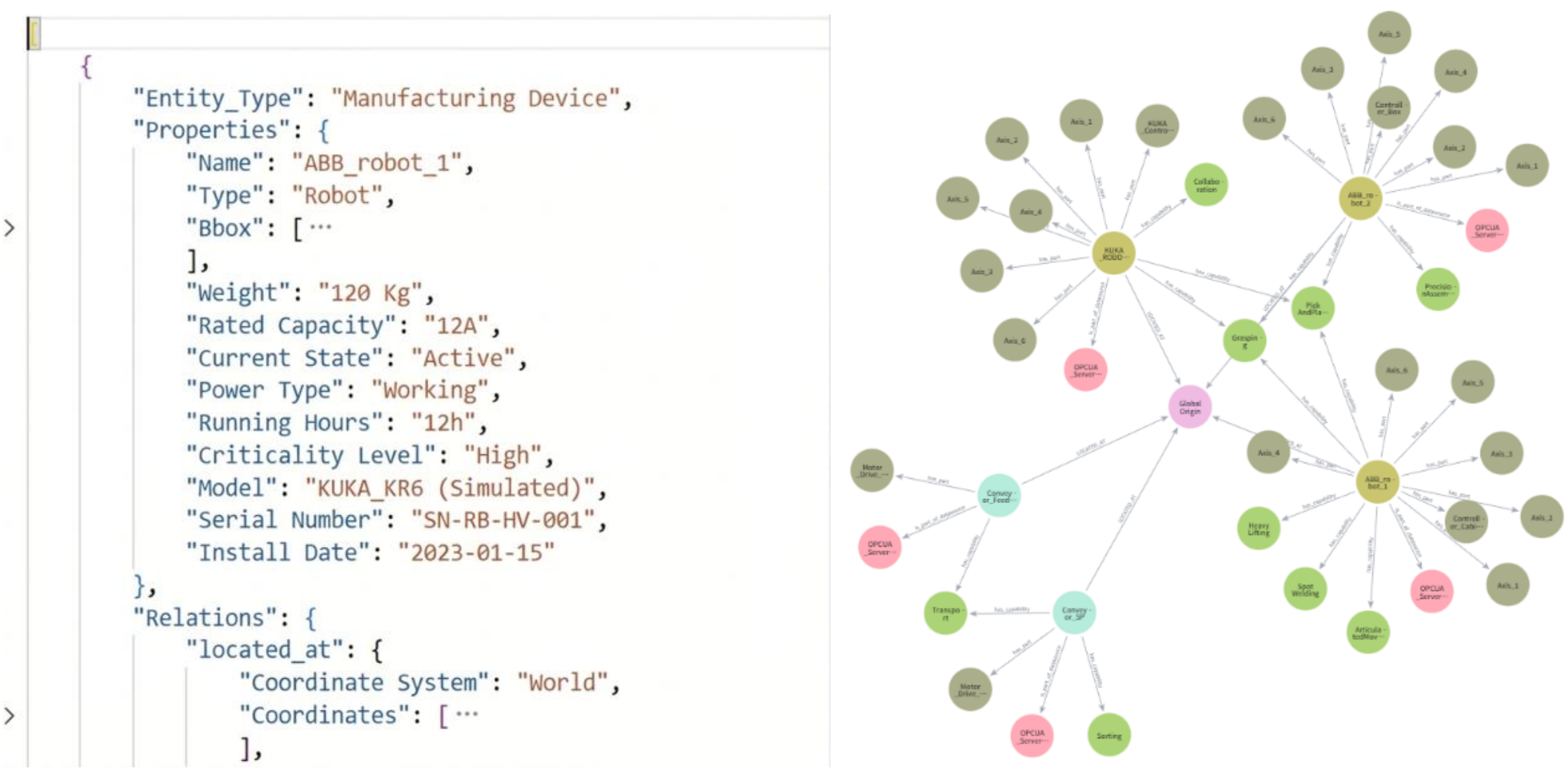

Figure 3. Semantic transformation workflow: From unstructured metadata (JSON vector) to structured Neo4j KG representation.

*Stage II Protocol Binding via MCP Tool-calling:* The reasoning core solves the "Heterogeneity Gap" by acting as a protocol mediator. The agent evaluates the Entity_Type and queries a capability mapping table. If Type == "Robot", it invokes an MCP tool (e.g., opcua_sync) to subscribe to 6-DOF joint states; if Type == "Conveyor", it prioritises linear velocity. This eliminates manual node-mapping, allowing the LLM to interact with hardware via standardised tool calls.

*Stage III HITL and Memory Persistent:* The final stage of the workflow synthesises all extracted data into a structured Knowledge Graph (KG) within Neo4j, which serves as the CDT's semantic foundation.

- Semantic Persistence: The finalised configuration serialises relationships between geometric STP entities, static RAG-extracted parameters, and dynamic OPC UA telemetry addresses.
- Real-time Synchronisation: The Three.js engine queries this KG to dynamically load the 3D scene and animate it with real-time telemetry, providing a high-fidelity monitoring interface.
- Closing the Feedback Loop: Moving beyond passive visualisation, the agent utilises bi-directional OPC UA write-nodes to push parameterised control signals back to the shop-floor. For instance, the system can autonomously adjust robot feed rates or initiate safety sequences based on the synchronised digital model's reasoning.
- HITL & Memory Persistence: To ensure scientific rigour, a HITL mechanism is triggered when perceptual confidence is low. These human corrections are stored in a persistent memory module, allowing the Agentic AI to learn from interventions and enhance the robustness of future autonomous actions.

### C. Experimental Results and Evaluation

*Perception Performance:* The YOLOv11 model was trained on 588 annotated images using an iterative "model-assisted + human correction" strategy. As shown in TABLE I and Figure 4, data augmentation significantly improved Precision and Recall.

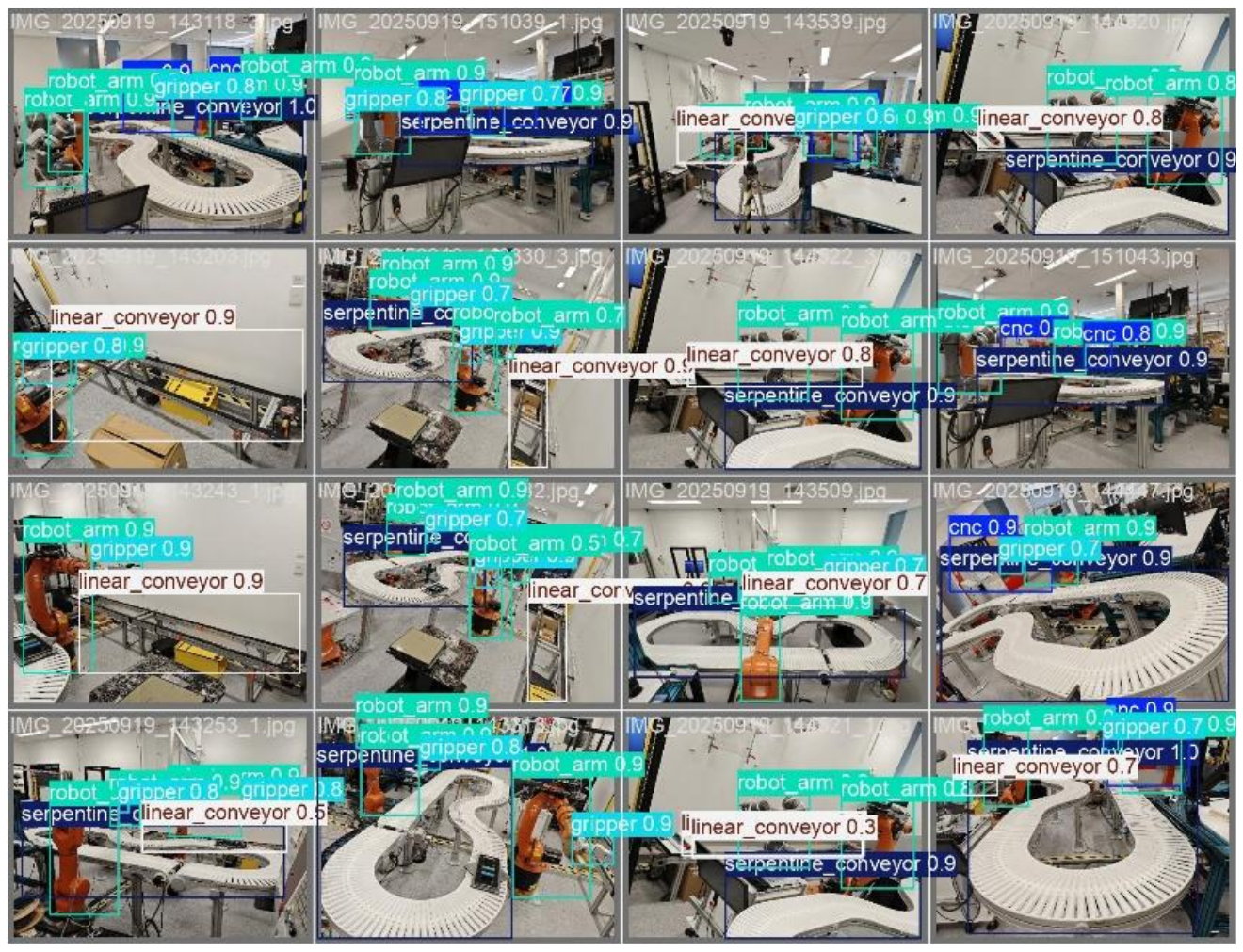

Figure 4. Visual verification of the perception module: Object detection and pose estimation in complex workshop environments.

*Qualitative Fidelity and Efficiency:* To evaluate the efficiency of the Agentic workflow, we compared it against the traditional manual CDT bespoke commissioning process (including 3D asset alignment, manual parameter entry from manuals, and OPC UA node binding).

- Manual Bespoke Commissioning: For a typical setup involving four major manufacturing entities, a skilled engineer requires several weeks to complete the data binding and spatial configuration.
- Agentic Workflow: The proposed autonomous pipeline, from raw camera input to a live Three.js

rendering, completes the cycle in an average of 2 hours.

The proposed autonomous pipeline reduced the deployment lead time by 95%, completing the cycle in an average of 2 hours compared to the multi-week duration required for manual commissioning. Qualitatively, the generated CDT achieves superior semantic fidelity; as illustrated in Figure 6, each virtual entity functions not merely as a geometric mesh (as shown in Figure 5) but as a knowledge-enriched node integrated with RAG-extracted parameters and live industrial data streams.

TABLE I. COMPARATIVE PERFORMANCE ANALYSIS OF YOLOV11 ACROSS DIVERSE TRAINING DATASETS AND AUGMENTATION STRATEGIES.

| Number of images | Precision | Recall | mAP@0.5 | mAP@0.5:0.95 |
|---|---|---|---|---|
| 370 | 0.898 | 0.936 | 0.933 | 0.732 |
| 588 | 0.952 | 0.95 | 0.965 | 0.737 |
| 588 + Data Aug. | 0.962 | 0.97 | 0.972 | 0.736 |

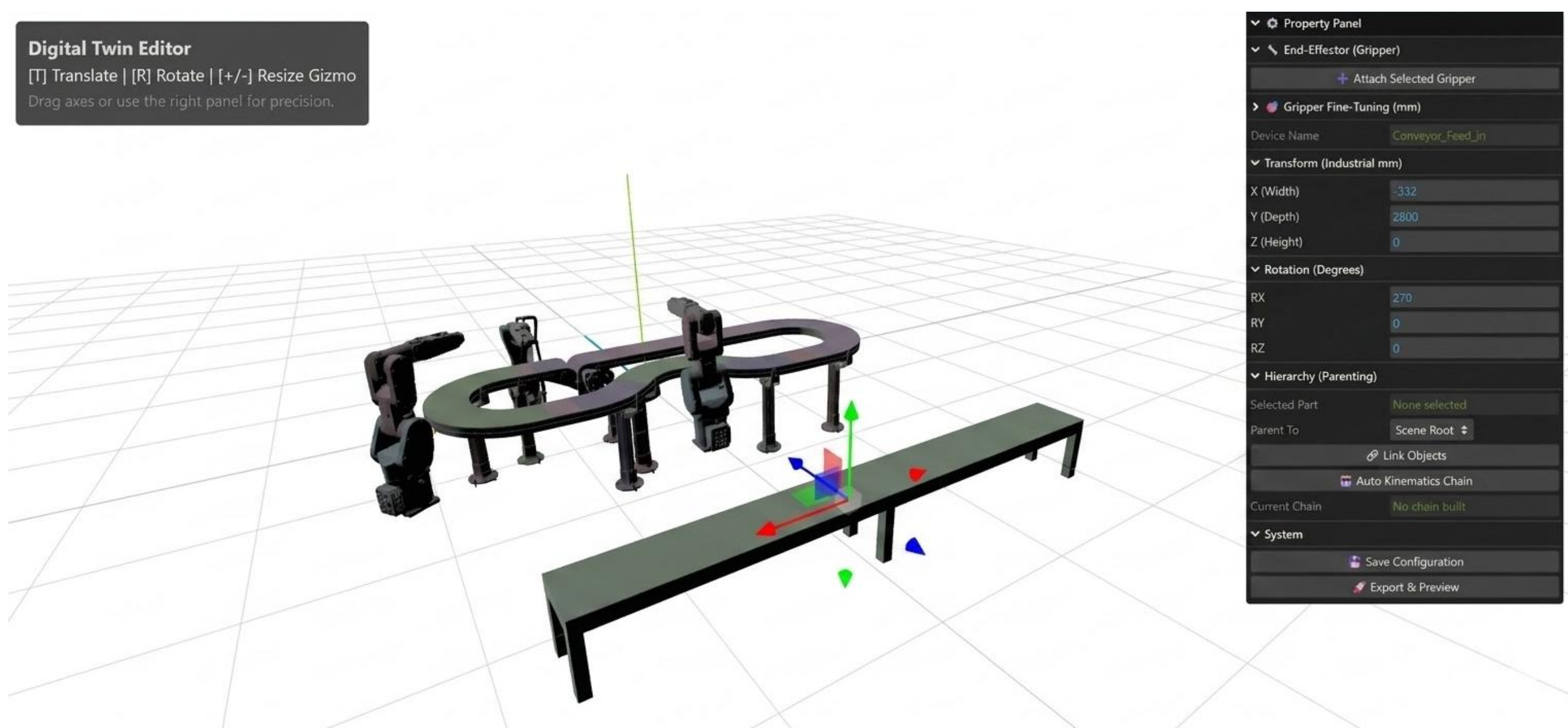

Figure 5. DT Editor interface: Spatial scene reconstruction and initial geometric-semantic mapping.

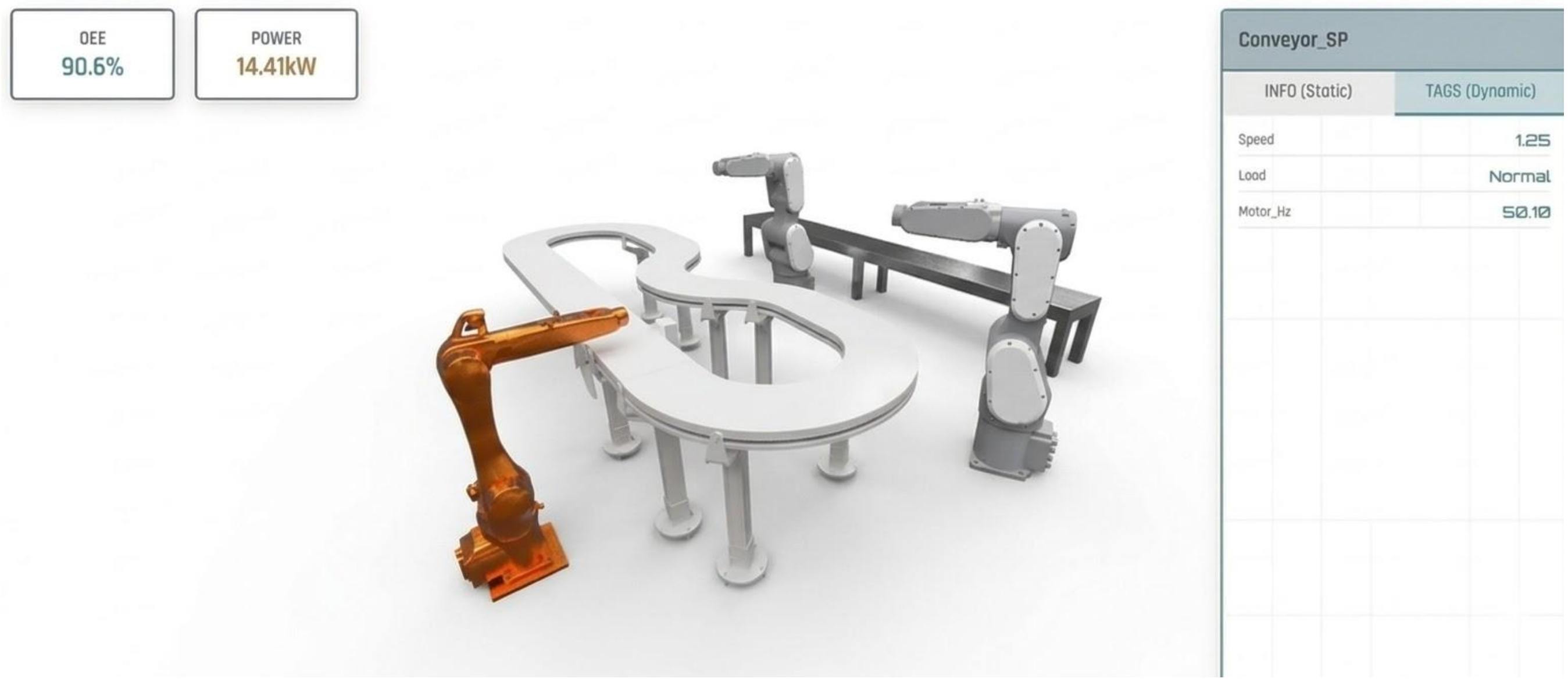

Figure 6. Live CDT dashboard: Real-time synchronization of industrial telemetry via MCP-enabled tool calls.

## IV. Conclusion & Future Work

This paper presents a systematic instantiation of a CDT framework driven by an Agentic workflow. By orchestrating a multi-agent system through LangGraph and leveraging the advanced reasoning capabilities of Large Language Models (LLMs), we have successfully bridged the gap between raw physical perception and high-fidelity, live-synchronised digital representations.

The integration of YOLOv11 for real-time spatial monitoring, combined with a dual-path semantic enrichment pipeline, utilising RAG for static technical parameters and the MCP for dynamic industrial protocol binding, enables an autonomous seamless interoperability commissioning of CDT. Experimental results in a robotic machining cell demonstrate that the system achieves not only high perceptual reliability (97.2% mAP@0.5) but also a significant 95% reduction in deployment lead time (shortening reconfiguration from approximately several weeks to several hours). This transition from manual, script-based configuration to autonomous orchestration proves both feasible and robust for modular Industry 5.0 environments.

Despite these advancements, limitations persist. The current RAG-based enrichment assumes digital documentation availability; future work will focus on Multimodal MLLMs to interpret legacy technical blueprints and unstructured hand-drawn diagrams. Furthermore, we intend to evolve the "Memory" module into a Distributed KG, facilitating cross-factory collective learning and experience sharing among decentralised Agentic AI systems.